\pdfoutput=1

\documentclass[11pt]{article}

\usepackage[final]{acl}

\usepackage{times}
\usepackage{latexsym}

\usepackage[T1]{fontenc}

\usepackage[utf8]{inputenc}

\usepackage{microtype}

\usepackage{inconsolata}

\usepackage{graphicx}
\usepackage{amsmath}
\usepackage[most]{tcolorbox}
\usepackage{multirow}
\usepackage{booktabs}
\usepackage{makecell}
\usepackage{pgfplots}
\usepackage{pgfplotstable}
\pgfplotsset{compat=1.17}
\usepackage{subcaption}
\usepackage{algorithm} 
\usepackage{algorithmic}
\usepackage{tikz}
\usetikzlibrary{calc}
\usepackage{float}  
\usepackage{tabularx}

\usepackage{etoolbox}
\usepackage{tikz}
\usepackage{xcolor}    
\usepackage{amsmath}   
\usepackage{wrapfig}

\title{Controlling Language Difficulty in Dialogues with Linguistic Features}

\author{Shuyao Xu, Wenguang Wang, Handong Gao,\\ {\bf Wei Kang},  {\bf Long Qin} \and {\bf Weizhi Wang} \\
Alibaba Group \\
\texttt{\{xushuyao.xsy, wangwenguang.wwg, handong.gaohd,\}@alibaba-inc.com} \\ \texttt{\{kw362539, ql362507, wangweizhi.wwz\}@alibaba-inc.com}
}

\begin{document}
\maketitle
\begin{abstract}
Large language models (LLMs) have emerged as powerful tools for supporting second language acquisition, particularly in simulating interactive dialogues for speaking practice. However, adapting the language difficulty of LLM-generated responses to match learners’ proficiency levels remains a challenge. This work addresses this issue by proposing a framework for  controlling language proficiency in educational dialogue systems. Our approach leverages three categories of linguistic features, readability features (e.g., Flesch-Kincaid Grade Level), syntactic features (e.g., syntactic tree depth), and lexical features (e.g., simple word ratio), to quantify and regulate text complexity. We demonstrate that training LLMs on linguistically annotated dialogue data enables precise modulation of language proficiency, outperforming prompt-based methods in both flexibility and stability. To evaluate this, we introduce Dilaprix, a novel metric integrating the aforementioned features, which shows strong correlation with expert judgments of language difficulty. Empirical results reveal that our approach achieves superior controllability of language proficiency while maintaining high dialogue quality. 
\end{abstract}

\section{Introduction}

Large language models (LLMs) have been widely adopted in the field of second language (L2) learning due to their ability to generate human-like text and engage users in interactive conversations. Recently, dialogue systems powered by LLMs, specifically designed to assist L2 learners in practicing speaking skills, are becoming increasingly common \cite{naseer2024chatbots, mahajan2022bela, young2023investigating, xu2024large}. These systems offer learners opportunities to engage in simulated conversations. However, for such systems to be truly effective, it is important to adapt the difficulty of the language used in these interactions to match the learner's proficiency level. This adaptation is particularly important for younger students, who may become disengaged or frustrated if the language is too complex.  Despite the remarkable performance of LLMs \cite{touvron2023llama, bai2023qwen}, achieving precise control over the language proficiency of LLM-generated utterances remains a challenge. 

To address the problem of accurately controlling the language proficiency level in utterances generated by LLMs, we investigate linguistic features strongly correlated with language difficulty. We leverage established readability formulas (e.g., Flesch-Kincaid Grade Level \cite{kincaid1975derivation}, the Gunning Fog Index \cite{gunning1969fog}, the Coleman-Liau Index \cite{coleman1975computer}). To capture deeper structural aspects, we incorporate syntactic complexity metrics derived from parsing trees \cite{ marcus-etal-1993-building, collins1997three}, analyzing grammatical structure. We further consider lexical features by analyzing word complexity aspects. By integrating these features, we aim to construct a framework for dynamically adjusting LLM output complexity to align with target L2 learner proficiency.

\begin{figure*}
    \centering
    \begin{tcolorbox}[colback=gray!1!white,colframe=black!75!black,arc=3mm]
    \small
\textbf{Context} \\
    Jane: \textit{Hello Lily, could you please come to the kitchen for a moment?} \\
    Lily: \textit{Of course.} \\

\begin{tabular}{lc}
\toprule
\textbf{Jane's responses under different language proficiency}  & \textbf{Dilaprix} \\
\midrule
\makecell[l]{\textit{Thank you for coming, Lily. Do you like meat?}} & 0.08 \\
\midrule
\makecell[l]{\textit{Thank you for coming, Lily. I appreciate your help in the kitchen. To start with, do you} \textit{like meat?}} & 0.30 \\
\midrule
\makecell[l]{\textit{Thank you for coming, Lily. I appreciate your help in the kitchen. To better understand your} \\ \textit{preferences, may I ask: do you like meat?}} & 0.55 \\
\midrule
\makecell[l]{\textit{ Ah, excellent, Lily, for you to grace us  with your presence in the kitchen. Now, to delve into a} \\ \textit{gastronomical inquiry: do you have an affinity for meat?}} & 0.81 \\ 
\bottomrule
\end{tabular}

\end{tcolorbox}
    \caption{Dialogue under different language proficiency.}
    \label{fig:dilaprix-example}
\end{figure*}

Evaluating complexity in dialogue presents unique difficulties, as standard models often depend on proficiency-labeled corpora, such as EFCAMDAT~\cite{geertzen2013automatic}, CLC-FCE~\cite{yannakoudakis-etal-2011-new}, Weebit~\cite{rama2021pretrainedtextrepresentationsuseful}, OneStopEnglish~\cite{vajjala-lucic-2018-onestopenglish}, Newsela~\cite{nushi2020newsela}   which are typically passages and unsuitable for conversations. Addressing this, we introduce the \textbf{Di}alogue \textbf{La}nguage \textbf{Pr}oficiency \textbf{I}nde\textbf{x} (Dilaprix). Dilaprix is designed as a comprehensive, interpretable metric calculated from the aforementioned linguistic features (readability, syntactic, lexical) to specifically quantify the complexity of dialogue utterances without reliance on pre-assigned proficiency levels.  Figure~\ref{fig:dilaprix-example} illustrates dialogue samples with varying Dilaprix.

Our contributions can be summarized as follows. First, we introduce Dilaprix, a comprehensive and interpretable metric designed to assess dialogue language proficiency. This metric is calculated based on readability, syntactic, and lexical features that are closely associated with language difficulty. Second, we propose leveraging these linguistic features to control the language difficulty of utterances generated by LLMs in dialogue systems. Through empirical evaluation, we demonstrate that an LLM trained on dialogue data annotated with the proposed linguistic features can effectively and accurately modulate language difficulty during conversations. This enables the model to generate responses that align with the learner's current proficiency level, thereby enhancing the adaptability and pedagogical utility of the dialogue system.

\section{Related Work}
Evaluating the language proficiency level of text is a well-researched problem. A common approach involves training regressors or classifiers using handcrafted linguistic features \cite{pilan-volodina-2018-investigating, pintard-francois-2020-combining}. More recently, researchers such as \citeauthor{deutsch-etal-2020-linguistic}, \citeauthor{arase-etal-2022-cefr}  and \citeauthor{kerz-etal-2021-automated} have leveraged deep learning models, including GRUs \cite{cho-etal-2014-properties} and Transformers \cite{vaswani2017attention}, to address this task. However, these methods are typically trained on labeled datasets, tailored to specific types of learning materials, such as essays and reading materials, which limits their transferability to new tasks, such as dialogues. 

In a recent study, \citeauthor{malik-etal-2024-tarzan} propose a novel approach to controlling LLMs for generating content aligned with specific Common European Framework of Reference (CEFR) language proficiency levels (A1–C2) \cite{council2001common}.  Their method utilizes CEFR level descriptions to guide the generation process. Experimental results indicate that while proprietary models like GPT-4 \cite{openai2024gpt4technicalreport} excel in prompting-based tasks, open-source models can achieve comparable performance through fine-tuning and reinforcement learning. Nevertheless, the inherent ambiguity of CEFR level definitions poses challenges for precise alignment, potentially limiting the reliability of such approaches.

\citeauthor{nguyen-etal-2024-multi} introduce a multi-objective fine-tuning framework to control linguistic properties in generated text. Their method demonstrates that integrating handcrafted linguistic features (e.g., the number of nouns, verbs, etc.) extracted by LFTK \cite{lee-lee-2023-lftk} into instruction-tuning workflows is an effective strategy for regulating the linguistic characteristics of content produced by LLMs. However, this work does not specifically aim to control language proficiency.

\section{Language difficulty related Linguistic Features}
\label{sec:ling-feats}
Readability scores are widely used metrics for evaluating text complexity. However, these scores, which usually rely on  syllable or character counts, are not always reliable indicators of text difficulty. To capture the multidimensional nature of language difficulty, we incorporate additional syntactic features and lexical features, providing a more comprehensive evaluation  and allowing for better control of language proficiency, encompassing both structural and lexical aspects of textual complexity.

\subsection{Readability Features}
We select four commonly used readability scores as features: Flesch Reading Ease, Flesch-Kincaid Grade Level, Gunning Fog Index, and Coleman-Liau Index~\cite{kincaid1975derivation, gunning1969fog, coleman1975computer}. 

\textbf{Flesch Reading Ease} ($F_{R}$) and \textbf{Flesch-Kincaid Grade Level} ($F_{G}$) are widely adopted readability measures that rely on average sentence length and the average number of syllables per word. Flesch Reading Ease evaluates the readability of a text by assigning a score, where higher values indicate greater ease of comprehension. Flesch-Kincaid Grade Level estimates the US grade level required for a reader to understand the text.

\textbf{Gunning Fog Index} ($G_F$), developed by Robert Gunning, is another widely used readability test designed to measure the difficulty of a text. It is calculated based on the average sentence length and the proportion of complex words — defined as words with three or more syllables. Higher scores indicate greater text complexity.

\textbf{Coleman-Liau Index} ($C_L$) is a readability metric that estimates the US grade level required for a reader to comprehend a text. Unlike the aforementioned readability scores,  the Coleman-Liau Index does not rely on syllable counting.  Instead, it relies on character counts per word and sentence length.

\subsection{Syntactic Features}
\label{sec:syntactic-features}
We incorporate five syntactic features in this work.

\textbf{Tree Depth} ($T_D$) is defined as the maximum depth of the syntactic trees across all sentences within the utterance.

\textbf{Leaf Node Count} ($L_N$) represents the highest number of leaf nodes found in any sentence within the utterance.

\textbf{Non-terminal Diversity} ($N_D$) measures the maximum number of unique non-terminal tags present in the syntactic trees of the sentences in the utterance.

\textbf{Subtree Complexity} ($S_C$) quantifies the maximum number of sub-trees identified in the syntactic trees of the sentences within the utterance.

\textbf{Utterance Length} ($U_L$) is the total number of tokens in the utterance. 

Among these, the first four features (Tree Depth, Leaf Node Count, Non-terminal Diversity, and Subtree Complexity) are extracted using a constituency parser ~\cite{klein-manning-2003-accurate} for evaluating and controlling dialogue language from the perspective of sentence structural complexity. Since an utterance may consist of one or more sentences, these features are computed at the utterance level by aggregating information from all sentences. 

\subsection{Lexical Features}
\label{sec:word-features}
We define two lexical features: Simple Word Ratio and Intermediate Word Ratio. These features are based on two expert-curated word lists commonly used in English language education. The simple word list comprises approximately 2,000 words typically taught in elementary English textbooks, while the intermediate word list expands this set to include an additional 4,000 words, covering vocabulary taught in both elementary and middle school English textbooks, for a total of 6,000 words.

\textbf{Simple Word Ratio} ($S_W$) is defined as the proportion of words in an utterance that are present in the simple word list.

\textbf{Intermediate Word Ratio} ($I_W$) is defined as the proportion of words in an utterance that are present in the intermediate word list.

\section{Method}

\subsection{Dialogue language proficiency Control}
We conduct our language proficiency control experiments on the textbook dialogue task, a specialized dialogue task designed for language learning. We describe the textbook dialogue in Appendix~\ref{sec:workflow-textbook-dialog}. Following \citeauthor{nguyen-etal-2024-multi}, language proficiency control through linguistic features can be modeled as a conditional instruction tuning problem. As shown in Figure~\ref{fig:prompt-example}, the textbook dialogue task is formulated as a dialogue completion problem, incorporating both dialogue task constraints and linguistic constraints.

\subsection{Datasets}
\label{sec:datasets}
 We constructed the dataset using user conversation data sourced from a textbook dialogue application. The user base of this application comprises students ranging from Grade 3 to Grade 9.  The data was initially split by topic into training (900 topics) and test (300 topics) sets. Approximately 18,000 training conversations (20 per topic) and 900 test conversations (3 per topic) were sampled.

\begin{figure}[H]
\begin{tcolorbox}[colback=gray!1!white,colframe=black!75!black,arc=3mm]
\scriptsize
\textbf{Instruction:} \par 
[flesch\_reading\_ease] for the Flesch-Kincaid Reading Ease;\par 
[flesch\_kincaid\_grade\_level] for Flesch-Kincaid Grade Level;\par 
[gunning\_fog] for the Gunning Fog Index;\par 
[coleman\_liau] for the Coleman Liau Index;\par 
[tree\_depth] The max Depth of the Constituency Parsing Trees of the sentences in your response;\par 
[leaf\_node\_count] The max number of leaf nodes of the Constituency Parsing Trees of the sentences in your response;\par 
[non\_terminal\_diversity] The max number of unique tags of the Constituency Parsing Trees of the sentences in your response;\par 
[subtree\_complexity] The max number of sub-trees of the Constituency Parsing Trees of the sentences in your response;\par 
[utterance\_length] the number of words in your response;\par 
[simple\_words\_ratio] the ratio of simple words in your response;\par 
[intermediate\_words\_ratio] the ratio of simple and intermediate words in your response.\par
\vspace{3 pt}
You are given a context and dialogue tasks, and are asked to play a role to continue the following conversation naturally. 

\vspace{3 pt} 
 
[DIALOGUE TASKS]\par 
1. Ask Anna if she can play the piano\par 
2. Ask Anna if she can ride a bike\par  
 \vspace{3 pt}

[CURRENT DIALOGUE TASK]\par 
2. Ask Anna if she can ride a bike\par 
 \vspace{3 pt} 
[CONTEXT]\par 
Ming:  Hi Anna, can you play the piano?\par 
Anna:  Yes, I can. \par 
 \vspace{3 pt}  
Your reply should consist of two parts:\par 
1. First part should respond to the user kindly based on the context;\par 
2. Second part should carry out the [CURRENT DIALOGUE TASK].\par 
Additionally, your response should abide by the following linguistic features:\par 
[flesch\_reading\_ease] 86.42 \par 
[flesch\_kincaid\_grade\_level] 3.07\par 
[gunning\_fog] 3.0\par 
[coleman\_liau] 2.99\par 
[tree\_depth] 9\par 
[leaf\_node\_count] 10\par 
[non\_terminal\_diversity] 14\par 
[subtree\_complexity] 22\par 
[utterance\_length] 18\par 
[simple\_words\_ratio] 0.8\par 
[intermediate\_words\_ratio] 1.0\par 
 \vspace{4 pt} 
\textbf{Response:} \par 
I am really looking forward to hearing you play. Can you ride a bike, Anna?
\end{tcolorbox}
\caption{A data example of language proficiency control in textbook dialogues. The LLMs are tasked with continuing a conversation based on a given context while completing the specified dialogue task. Additionally, the generated responses are expected to adhere as closely as possible to predefined linguistic feature constraints. } 
\label{fig:prompt-example}
\end{figure}

 A conversation is composed of multiple turns where user utterances and AI responses alternate.  We randomly select one turn from the conversation. The dialogue content prior to the AI's response in this sampled turn serves as the conversational context. To generate training data for controlling linguistic features, we require utterances  with varying linguistic feature configurations. To this end, we employed an LLM (\textit{Qwen-plus}\footnote{https://www.alibabacloud.com}), to produce utterances conditioned on the conversational context and instructions specifying one of the six CEFR proficiency levels, as outlined by \citeauthor{malik-etal-2024-tarzan}. From each generated utterance, we extracted 11 linguistic features (as described in Section~\ref{sec:ling-feats}).  Each training instance was then formed by pairing the original context with these extracted linguistic features. This process ensured that each dialogue context was associated with six distinct responses, each reflecting a unique linguistic feature configuration.  In total, this approach generated approximately 100,000 instances for instruction tuning.

Reinforcement learning with human feedback (RLHF) has been demonstrated an effective way to align output with human preference \cite{christiano2017deep, bai2022training, deepseekai2025deepseekr1incentivizingreasoningcapability}. In this work, we optimize the dialogue language proficiency control model using Direct Preference Optimization (DPO) \cite{rafailov2023direct}. To construct the DPO dataset, we sampled 30,000 instruction-response pairs from the instruction-tuning set, ensuring that each pair consisted of an instruction and its preferred response. For the negative responses, we generated two types of negative samples: (1) responses that were contextually correct with respect to the dialogue context and task but exhibited incorrect linguistic features, and (2) responses that were either inappropriate for the dialogue context or failed to fulfill the specified dialogue task.

\subsection{Models}

We conduct dialogue language linguistic control experiments on Qwen2.5-7B-Instruct-1M~\cite{yang2025qwen251mtechnicalreport} and LLaMA3.1-8B-Instruct~\cite{grattafiori2024llama3herdmodels} with LLaMA-Factory~\cite{zheng2024llamafactory}.

\subsection{Baselines}
Two baseline methods are employed for comparison. The first is a prompt-based approach utilizing the \textit{Qwen-plus}, selected for its favorable balance between cost and dialogue quality. This method incorporates prompts guided by CEFR language proficiency levels \cite{malik-etal-2024-tarzan}. The second baseline employs instruction tuning, also guided by CEFR language proficiency levels. Similar to the dataset construction described in Section~\ref{sec:datasets}, the instruction tuning data for this baseline is created by replacing linguistic features with their corresponding CEFR level descriptions.

\section{Metrics}
The dialogue model evaluation focuses on two aspects: dialogue quality and the language proficiency controllability.

\subsection{Response Success Rate}
In textbook dialogue scenarios, LLMs are expected to respond naturally while simultaneously fulfilling  a specified dialogue task. Consequently, a response is considered successful only if it satisfies both requirements.  We measure dialogue quality of dialogue models using the Response Success Rate (RSR), defined as the ratio of successful responses to the total number of responses.  

Leveraging strong LLMs to automate dialogue assessment has become a popular approach \cite{zheng2023judgingllmasajudgemtbenchchatbot}. In our work, we employ a strong LLM, \textit{Qwen-max}\footnote{https://chat.qwen.ai}, to evaluate whether a given response successfully meets these criteria based on the context and dialogue task. The performance of the dialogue evaluator is presented in Appendix~\ref{sec:dialog-assess}. 

\subsection{Dialogue Language Proficiency Index}
\label{sec:dilaprix}
Handcrafted features are commonly employed for predicting language proficiency levels \cite{pilan-volodina-2018-investigating, pintard-francois-2020-combining}. However, these approaches often require significant expert effort in data annotation. By utilizing  a curated set of linguistic features, as detailed in Section~\ref{sec:ling-feats}, we demonstrate that a simple aggregation of these features enables  a comprehensive and accurate evaluation of dialogue language proficiency.  The Dilaprix is formally defined as follows:

\begin{align}
\text{Dilaprix} = \frac{1}{|\mathcal{X}|}\sum_{x_i}^{x_i\in \mathcal{X}}\tau(x_i)
\end{align}

\begin{align}
\tau(x_i)=
\begin{cases}
1 - \text{clamp}(\frac{x_i - \alpha_i}{\beta_i - \alpha_i}, 0, 1) & \text{if}\  x_i \in \mathcal{X}^{\prime} \\
\text{clamp}(\frac{x_i - \alpha_i}{\beta_i - \alpha_i}, 0, 1)& \text{otherwise}
\end{cases}
\end{align}

where, $\mathcal{X} =$ \{$F_R$, $F_G$, $G_F$, $C_L$, $T_D$, $L_N$, $N_D$, $S_C$, $U_L$, $S_W$, $I_W$\}; $\mathcal{X}^{\prime}=$\{$F_R$, $S_W$, $I_W$\}, is the subset of features that are inversely related to difficulty; and $\alpha_i$ and $\beta_i$ denote the 5th and 95th percentiles, respectively, of the feature $x_i$, as empirically determined from the textbook dialogue corpus. These percentiles are chosen to exclude outliers and ensure robust normalization. $\text{clamp}(v, 0, 1)$ constrains the value $v$ within the interval $[0, 1]$.

\begin{table*}[htbp]
\centering
\scriptsize
\begin{tabular}{l|l|cccccc}
\toprule
\textbf{Method} & \textbf{Metrics} & \textbf{A1} & \textbf{A2} & \textbf{B1} & \textbf{B2} & \textbf{C1} & \textbf{C2} \\ 
\midrule
\multirow{3}{2.cm}{Prompt-Based} 
    & RSR       & 0.932 & 0.959 & 0.917 & 0.935 & 0.913 & 0.790 \\ 
    \cmidrule{2-8}
    & Dilaprix  & \makecell{$0.224$  $\pm 0.120$} 
                & \makecell{$0.231$  $\pm 0.126$} 
                & \makecell{$0.393$  $\pm 0.153$} 
                & \makecell{$0.494$  $\pm 0.167$} 
                & \makecell{$0.524$  $\pm 0.167$} 
                & \makecell{$0.685$  $\pm 0.187$} \\
\midrule
\multirow{3}{2.cm}{LLAMA-CEFR} 
    & RSR       & 0.957 & 0.957 & 0.965 & 0.973 & 0.970 & 0.958 \\ 
    \cmidrule{2-8}
    & Dilaprix  & \makecell{$0.211$  $\pm 0.116$} 
                & \makecell{$0.217$  $\pm 0.127$} 
                & \makecell{$0.389$  $\pm 0.154$} 
                & \makecell{$0.480$  $\pm 0.173$} 
                & \makecell{$0.509$  $\pm 0.171$} 
                & \makecell{$0.673$  $\pm 0.191$} \\
\midrule
\multirow{3}{2.cm}{QWEN-CEFR} 
    & RSR       & 0.957 & 0.974 & 0.975 & 0.969 & 0.967 & 0.954 \\ 
    \cmidrule{2-8}
    & Dilaprix  & \makecell{$0.210$  $\pm 0.119$} 
                & \makecell{$0.218$  $\pm 0.124$} 
                & \makecell{$0.379$  $\pm 0.154$} 
                & \makecell{$0.482$  $\pm 0.169$} 
                & \makecell{$0.512$  $\pm 0.170$} 
                & \makecell{$0.679$  $\pm 0.194$} \\

\bottomrule
\end{tabular}
\caption{Results of prompt-based baseline and CEFR instruction tuning baselines on the textbook dialogue test set. Prompt-Based denotes the prompt-based baseline, LLAMA-CEFR refers to the performance of Llama3.1-8B-Instruct after being fine-tuned on the CEFR-instructed textbook dialogue dataset. Similarly, QWEN-CEFR represents the results of Qwen2.5-7B-Instruct trained on the same dataset. Dilaprix (mean $\pm$ standard deviation) is the averaged across instances in the test set.  }.
\label{tab:baseline-results}
\end{table*}

\begin{table*}[h]
\centering
\scriptsize
\begin{tabular}{l|l|cccccc}
\toprule
Method & \textbf{Metrics} & $t=0.0$ & $t=0.2$ & $t=0.4$ & $t=0.6$ & $t=0.8$ & $t=1.0$ \\ 

\midrule
\multirow{3}{2cm}{LLAMA-FT} 
    & RSR       & 0.806 & 0.944 & 0.944 & 0.966 & 0.962 & 0.949 \\ 
    \cmidrule{2-8}
    & Dilaprix  & \makecell{$0.096$  $\pm 0.074$} 
                & \makecell{$0.218$  $\pm 0.093$} 
                & \makecell{$0.357$  $\pm 0.105$} 
                & \makecell{$0.547$  $\pm 0.104$} 
                & \makecell{$0.685$  $\pm 0.106$} 
                & \makecell{$0.834$  $\pm 0.086$} \\
\midrule
\multirow{3}{2cm}{LLAMA-DPO} 
    & RSR       & 0.863 & 0.946 & 0.975 & 0.991 & 0.977 & 0.953 \\ 
    \cmidrule{2-8}
    & Dilaprix  & \makecell{$0.080$  $\pm 0.063$} 
                & \makecell{$0.186$  $\pm 0.084$} 
                & \makecell{$0.356$  $\pm 0.096$} 
                & \makecell{$0.549$  $\pm 0.094$} 
                & \makecell{$0.695$  $\pm 0.094$} 
                & \makecell{$0.846$  $\pm 0.078$} \\

\midrule
\multirow{3}{2.cm}{QWEN-FT} 
    & RSR       & 0.868 & 0.941 & 0.960 & 0.970 & 0.945 & 0.928 \\ 
    \cmidrule{2-8}
    & Dilaprix  & \makecell{$0.099$  $\pm 0.071$} 
                & \makecell{$0.216$  $\pm 0.083$} 
                & \makecell{$0.363$  $\pm 0.099$} 
                & \makecell{$0.546$  $\pm 0.103$} 
                & \makecell{$0.713$  $\pm 0.103$} 
                & \makecell{$0.840$  $\pm 0.088$} \\
\midrule
\multirow{3}{2cm}{QWEN-DPO} 
    & RSR       & 0.872 & 0.953 & 0.990 & 0.977 & 0.977 & 0.963 \\ 
    \cmidrule{2-8}
    & Dilaprix  & \makecell{$0.073$  $\pm 0.057$} 
                & \makecell{$0.192$  $\pm 0.083$} 
                & \makecell{$0.361$  $\pm 0.094$} 
                & \makecell{$0.565$  $\pm 0.099$} 
                & \makecell{$0.752$  $\pm 0.094$} 
                & \makecell{$0.883$  $\pm 0.070$} \\

\bottomrule
\end{tabular}
\caption{Results of linguistic feature trained textbook dialogue models. Models are evaluated based on groups of linguistic features $x_i\in \mathcal{X}$ sharing a common value $t=\tau(x_i)$, for $t\in \{0.0, 0.2, 0.4, 0.6, 0.8, 1.0\}$. LLAMA-FT refers to the model fine-tuned on a training set comprising textbook dialogues with linguistic features. LLAMA-DPO represents the version further optimized using DPO. Similarly, QWEN-FT denotes the model fine-tuned on the same training set, and QWEN-DPO refers to its DPO-optimized version.}
\label{tab:sft-dpo-results}
\end{table*}

We further investigate the alignment between Dilaprix and expert judgments regarding the difficulty of utterances from textbook dialogues. A sample of 100 utterances was selected. As assigning absolute difficulty scores to individual utterances is challenging, we employed a pairwise comparison method. This process generated 4,950 utterance pairs from the 100 samples. Three experts were independently tasked with identifying the more difficult utterance within each pair. An utterance's difficulty score was calculated as its "winning rate" — the proportion of pairs in which it was judged more difficult. The final ground truth difficulty scores were obtained by averaging the scores assigned by the three experts. 

Dilaprix demonstrated a strong correlation with ground truth difficulty scores, achieving a Pearson correlation coefficient (PCC) of 0.950. This surpasses the average PCC among the experts (0.935), demonstrating that Dilaprix is a reliable metric for assessing language difficulty in dialogues.

\section{Evaluation}
\subsection{Prompt-based and CEFR instruction finetuned baselines}

\begin{figure*}[h]
\centering
\footnotesize

\begin{subfigure}[b]{0.45\textwidth} 
\centering
\begin{tikzpicture}
\begin{axis}[
    xlabel={Dilaprix},
    ylabel={RSR},
    xmin=0, xmax=1,
    ymin=0.78, ymax=1,
    xtick={0,0.1,...,1},
    ytick={0.75,0.80,...,1},
    legend pos=south east,
    ymajorgrids=true,
    grid style=dashed,
    width=1.1\textwidth, 
    height=0.7\textwidth,
    legend style={font=\small, scale=0.8},
]

\addplot[
    color=red,
    only marks,
    mark=*,
] coordinates {
(0.096,0.806) (0.165,0.903) (0.218,0.944) (0.28,0.934) (0.357,0.944) (0.461,0.969) (0.547,0.966) (0.617,0.978) (0.685,0.962) (0.754,0.951) (0.834,0.949)
};
\addlegendentry{LLAMA-FT}

\addplot[
    color=blue,
    only marks,
    mark=*,
] coordinates {
(0.08,0.863) (0.126,0.919) (0.186,0.946) (0.263,0.97) (0.355,0.975) (0.464,0.98) (0.549,0.991) (0.63,0.98) (0.695,0.977) (0.767,0.974) (0.846,0.953)
};
\addlegendentry{LLAMA-DPO}

\addplot[no markers, blue, dashed,  mark=o, domain=0.08:0.846, samples=100] { -28.5878 * x^6 + 86.5285 * x^5 -105.3992 * x^4 + 65.8943 * x^3 -22.4753 * x^2 + 4.0945 * x + 0.65 };

\addplot[no markers, red, dashed, mark=o, domain=0.096:0.834, samples=100] {  -29.3707 * x^6 + 111.0752 * x^5 -161.8532 * x^4 + 116.1158 * x^3 -43.3248 * x^2 + 8.1051 * x + 0.3365 };

\end{axis}
\end{tikzpicture}
\caption{}
\label{fig:dilaprix-rsr-line-right} 
\end{subfigure}
\hspace{0.25cm}
\hfill 
\begin{subfigure}[b]{0.45\textwidth} 
\centering
\begin{tikzpicture}
\begin{axis}[
    xlabel={Dilaprix},
    xmin=0, xmax=1,
    ymin=0.78, ymax=1,
    xtick={0,0.1,...,1},
    ytick={0.75,0.80,...,1},
    legend pos=south east,
    ymajorgrids=true,
    grid style=dashed,
    width=1.1\textwidth, 
    height=0.7\textwidth,
    legend style={font=\small, scale=0.8},
]

\addplot[
    color=red,
    only marks,
    mark=*,
] coordinates {
(0.099,0.868) (0.16,0.907) (0.216,0.941) (0.281,0.959) (0.363,0.96) (0.462,0.974) (0.546,0.97) (0.633,0.957) (0.713,0.945) (0.787,0.934) (0.84,0.928)
};
\addlegendentry{QWEN-FT}

\addplot[
    color=blue,
    only marks,
    mark=*,
] coordinates {
(0.073,0.872) (0.128,0.913) (0.192,0.953) (0.267,0.973) (0.361,0.99) (0.478,0.987) (0.565,0.977) (0.658,0.984) (0.752,0.977) (0.829,0.975) (0.883,0.963)
};
\addlegendentry{QWEN-DPO}

\addplot[no markers, blue, dashed,  mark=o, domain=0.073:0.883, samples=100] { -1.9422 * x^4 + 4.5878 * x^3 -4.0155 * x^2 + 1.5158 * x + 0.7791 };

\addplot[no markers, red, dashed, mark=o, domain=0.099:0.84, samples=100] { 5.2501 * x^6 -9.6402 * x^5 + 3.3765 * x^4 + 3.9777 * x^3 -4.2891 * x^2 + 1.6178 * x + 0.7447  };

\end{axis}
\end{tikzpicture}
\caption{}
\label{fig:fig:dilaprix-rsr-line-left} 
\end{subfigure}

\caption{Dialogue quality (RSR) achieved versus language proficiency (Dilaprix). Comparisons are presented for (a) LLAMA and (b) QWEN architectures, illustrating performance across the controllable complexity range.}
\label{fig:dilaprix-rsr-line} 
\end{figure*}

We first conducted language proficiency control experiments using a prompt-based method, implementing a few-shot prompting strategy. As shown in Table \ref{tab:baseline-results}, the prompt-based approach exhibits a significant decline in dialogue quality at the C2 proficiency level. In contrast, models employing a CEFR instruction fine-tuning method (with 7B or 8B parameters) achieve comparable language proficiency control while demonstrating superior dialogue quality. Specifically, the fine-tuned Qwen model achieves an average RSR of 0.966 across all six CEFR levels, significantly outperforming the prompt-based method, which yields an average RSR of 0.908.

\subsection{Evaluation of Linguistic Feature Controlled Dialogue Models}

As shown in Figure~\ref{fig:prompt-example}, during inference, linguistic features must be explicitly integrated into the prompt. In this work, each feature is assigned a value $x_i$ that satisfies $\tau(x_i)=t$, with all linguistic features sharing the same parameter $t$.

As shown in Table~\ref{tab:sft-dpo-results}, dialogue models controlled by linguistic features demonstrate effective controllability over language proficiency, measured by the Dilaprix. These models achieve a wide range of Dilaprix  (e.g., QWEN-DPO spans 0.073 to 0.883). In contrast, a comparable model guided by CEFR level instructions yielded only six distinct complexity levels as shown in Table~\ref{tab:baseline-results}, corresponding to a narrower Dilaprix range (minimum 0.211, maximum 0.685).  Moreover, the feature-controlled models generate utterances with markedly more consistent complexity, reflected in a substantially lower Dilaprix standard deviation (STD) (QWEN-DPO: 0.083 compared to 0.155 for QWEN-CEFR). This lower Dilaprix STD indicates greater stability in the linguistic complexity of the generated utterances.

Figure~\ref{fig:dilaprix-rsr-line} presents the RSR achieved across the controllable Dilaprix range for the QWEN and LLAMA models. Dashed lines represent fitted trend curves, obtained by fitting polynomial functions. The degree of these polynomials was determined either when the coefficient of determination ($R^2$) exceeded 0.98 or when the maximum polynomial degree of 6 was reached.  To provide a comprehensive assessment of model performance concerning language proficiency controllability and dialogue quality, we introduce the Area Under the Curve (AUC) metric. This metric captures the overall trend shown in Figure~\ref{fig:dilaprix-rsr-line}, where RSR varies with Dilaprix, by calculating the integral of the RSR curve across the operational range of Dilaprix supported by the models. 

We observe lower RSR under low Dilaprix conditions.  This phenomenon can likely be attributed to the reason  that textbook dialogues  require the model to perform specific dialogue tasks, which may involve complex language structures or lengthy sentences. For instance, a task such as "Say you like drawing mountains because they look beautiful and ask Tom what he thinks about drawing at the park" necessitates the use of long utterances, which may conflict with the constraints imposed by a low Dilaprix. Given that response success is judged on both natural interaction and the accurate fulfillment of the specified dialogue task, this inherent contradiction likely impedes the model's ability to meet both criteria, thus lowering the RSR. 

According to Table~\ref{tab:auc-results}, applying DPO  yields consistent performance improvements for both QWEN and LLAMA models. Enhanced controllability over language proficiency, coupled with the maintenance of good dialogue quality, is evidenced by wider Dilaprix ranges (DR) and larger AUC. Moreover, the observed reduction in the average standard deviation (STD) of the Dilaprix metric indicates increased stability in controlling language proficiency. The QWEN-DPO model exhibited the strongest performance overall.

\begin{table}[h]
\small
\centering
\begin{tabular}{l|ccc}
\toprule
\textbf{Method} & \textbf{AUC}\,($\uparrow$) & \textbf{ DR}\,($\uparrow$) & \textbf{ STD}\,($\downarrow$) \\ 
\midrule
LLAMA-FT & 0.698 & 0.739 & 0.097 \\ 

LLAMA-DPO & 0.741 & 0.766 & 0.086 \\ 

\midrule
QWEN-FT & 0.703 & 0.741 & 0.093 \\ 

QWEN-DPO & \textbf{0.786} & \textbf{0.810} & \textbf{0.084} \\ 

\bottomrule
\end{tabular}
\caption{Performance comparison of language proficiency-controlled models based on QWEN and LLAMA.}.
\label{tab:auc-results}
\end{table}

\begin{table}[h]
\small
\centering
\begin{tabular}{l|ccc}
\toprule
& \textbf{AUC}\,($\uparrow$) & \textbf{ DR}\,($\uparrow$) & \textbf{ STD}\,($\downarrow$) \\ 
\midrule
BASELINE & 0.703 & 0.741 & 0.093 \\ 
\midrule
- $F_R$ & 0.679 & 0.710 & 0.094 \\ 

- $F_G$ & 0.692 & 0.722 & 0.094 \\ 

- $G_F$ & 0.690 & 0.725 & 0.091 \\ 

- $C_L$ & 0.680 & 0.713 & 0.099 \\
\midrule
- $T_D$ & 0.685 & 0.720 & 0.094 \\

- $L_N$ & 0.679 & 0.712 & 0.093 \\

- $N_D$ & 0.666 & 0.695 & 0.094 \\

- $S_C$ & 0.678 & 0.716 & 0.095 \\

- $U_L$ & 0.724 & 0.748 & 0.090 \\
\midrule
- $S_W$ & 0.647 & 0.676 & 0.092 \\

- $I_W$ & 0.687 & 0.718 & 0.096 \\
\bottomrule
\end{tabular}
\caption{Ablation study examining the impact of removing individual linguistic features on model performance.}
\label{tab:ablation-study-on-single-feature}
\end{table}

\subsection{Linguistic Feature Controllability}

\begin{figure*}[h]
\footnotesize
    \centering

\begin{subfigure}[b]{0.45\textwidth} 
\centering
\begin{tikzpicture}
\def\centerx{0}
\def\centery{0}
\def\radius{2}
\def\angles{0, 32.73, 65.46, 98.19, 130.92, 163.65, 196.38, 229.11, 261.84, 294.57, 327.30}
\def\anglesmat{{\angles}}

\foreach \angle in \angles {
    \draw[gray!50] (\centerx,\centery) -- ++(\angle:\radius);
}
\foreach \r/\label in {0.2/0.2, 0.4/0.4, 0.6/0.6, 0.8/0.8, 1.0/1.0} {
    \draw[gray!50] (\centerx,\centery) circle (\radius*\r);
    \node[anchor=west] at (90:\radius*\r) {\label};
}

\def\datarowA{{0.2365, 0.0537, 0.0132, 0.0345, 0.0262, 0.0282, 0.1027, 0.0503, 0.0016, 0.1857, 0.0923}}
\def\datarowB{{0.343, 0.1667, 0.1226, 0.2613, 0.2175, 0.1482, 0.3, 0.1809, 0.1969, 0.2771, 0.2077}}
\def\datarowC{{0.4811, 0.29, 0.2705, 0.4339, 0.3838, 0.26, 0.4391, 0.3081, 0.3569, 0.4429, 0.3538}}
\def\datarowD{{0.64, 0.4612, 0.4347, 0.6331, 0.5562, 0.4959, 0.6836, 0.5478, 0.5909, 0.5914, 0.4923}}
\def\datarowE{{0.7672, 0.597, 0.5633, 0.7347, 0.6625, 0.6894, 0.8364, 0.745, 0.7878, 0.7143, 0.6385}}
\def\datarowF{{0.947, 0.7751, 0.7173, 0.8688, 0.75, 0.9153, 0.9845, 0.9616, 1.0, 0.88, 0.9615}}

\def\coleur{blue}
\foreach \idx in {0, 1, ..., 10} {
    \pgfmathsetmacro{\angle}{\anglesmat[\idx]}   
    \pgfmathsetmacro{\datapoint}{\datarowA[\idx]} 
    \coordinate (\idx) at (\angle:\radius*\datapoint); 
}
\draw[thick,\coleur] (0) -- (1) -- (2) -- (3) -- (4) -- (5) -- (6) -- (7) -- (8) -- (9) -- (10) -- cycle;
\foreach \point in {0, 1, ..., 10}  {
    \fill[\coleur] (\point) circle (2pt);
}

\def\coleur{brown}
\foreach \idx in {0, 1, ..., 10} {
    \pgfmathsetmacro{\angle}{\anglesmat[\idx]}   
    \pgfmathsetmacro{\datapoint}{\datarowB[\idx]} 
    \coordinate (\idx) at (\angle:\radius*\datapoint); 
}
\draw[thick,\coleur] (0) -- (1) -- (2) -- (3) -- (4) -- (5) -- (6) -- (7) -- (8) -- (9) -- (10) -- cycle;
\foreach \point in {0, 1, ..., 10}  {
    \fill[\coleur] (\point) circle (2pt);
}

\def\coleur{purple}
\foreach \idx in {0, 1, ..., 10} {
    \pgfmathsetmacro{\angle}{\anglesmat[\idx]}   
    \pgfmathsetmacro{\datapoint}{\datarowC[\idx]} 
    \coordinate (\idx) at (\angle:\radius*\datapoint); 
}
\draw[thick,\coleur] (0) -- (1) -- (2) -- (3) -- (4) -- (5) -- (6) -- (7) -- (8) -- (9) -- (10) -- cycle;
\foreach \point in {0, 1, ..., 10}  {
    \fill[\coleur] (\point) circle (2pt);
}

\def\coleur{yellow}
\foreach \idx in {0, 1, ..., 10} {
    \pgfmathsetmacro{\angle}{\anglesmat[\idx]}   
    \pgfmathsetmacro{\datapoint}{\datarowD[\idx]} 
    \coordinate (\idx) at (\angle:\radius*\datapoint); 
}
\draw[thick,\coleur] (0) -- (1) -- (2) -- (3) -- (4) -- (5) -- (6) -- (7) -- (8) -- (9) -- (10) -- cycle;
\foreach \point in {0, 1, ..., 10}  {
    \fill[\coleur] (\point) circle (2pt);
}

\def\coleur{orange}
\foreach \idx in {0, 1, ..., 10} {
    \pgfmathsetmacro{\angle}{\anglesmat[\idx]}   
    \pgfmathsetmacro{\datapoint}{\datarowE[\idx]} 
    \coordinate (\idx) at (\angle:\radius*\datapoint); 
}
\draw[thick,\coleur] (0) -- (1) -- (2) -- (3) -- (4) -- (5) -- (6) -- (7) -- (8) -- (9) -- (10) -- cycle;
\foreach \point in {0, 1, ..., 10}  {
    \fill[\coleur] (\point) circle (2pt);
}

\def\coleur{red}
\foreach \idx in {0, 1, ..., 10} {
    \pgfmathsetmacro{\angle}{\anglesmat[\idx]}   
    \pgfmathsetmacro{\datapoint}{\datarowF[\idx]} 
    \coordinate (\idx) at (\angle:\radius*\datapoint); 
}
\draw[thick,\coleur] (0) -- (1) -- (2) -- (3) -- (4) -- (5) -- (6) -- (7) -- (8) -- (9) -- (10) -- cycle;
\foreach \point in {0, 1, ..., 10}  {
    \fill[\coleur] (\point) circle (2pt);
}

\foreach \angle/\label in {0/$F_R$, 32.73/$F_G$, 65.46/$G_F$, 98.19/$C_L$, 130.92/$T_D$, 163.65/$L_N$, 196.38/$N_D$, 229.11/$S_C$, 261.84/$U_L$, 294.57/$S_W$, 327.30/$I_W$} {
    \node at (\angle:\radius*1.12) {\label};
}
\end{tikzpicture}
\caption{LLAMA-FT}
\end{subfigure}
\hfill
\begin{subfigure}[b]{0.45\textwidth} 
\centering
\begin{tikzpicture}
\def\centerx{0}
\def\centery{0}
\def\radius{2}
\def\angles{0, 32.73, 65.46, 98.19, 130.92, 163.65, 196.38, 229.11, 261.84, 294.57, 327.30}
\def\anglesmat{{\angles}}

\def\datarowA{{0.1451, 0.0, 0.0, 0.0, 0.0188, 0.0135, 0.1, 0.0406, 0.0116, 0.1629, 0.0846}}
\def\datarowB{{0.2643, 0.1039, 0.1185, 0.1795, 0.2238, 0.1224, 0.2564, 0.1666, 0.195, 0.22, 0.1846}}
\def\datarowC{{0.4548, 0.2683, 0.2847, 0.4122, 0.4312, 0.2494, 0.4018, 0.3125, 0.37, 0.4371, 0.3923}}
\def\datarowD{{0.6348, 0.4578, 0.4661, 0.6292, 0.6025, 0.4724, 0.6627, 0.5328, 0.5784, 0.5886, 0.5231}}
\def\datarowE{{0.7805, 0.6164, 0.6079, 0.7446, 0.675, 0.6641, 0.8318, 0.7222, 0.765, 0.72, 0.6846}}
\def\datarowF{{0.9766, 0.8116, 0.7771, 0.8826, 0.7512, 0.8647, 0.9655, 0.9038, 1.0, 0.88, 1.0}}

\foreach \angle in \angles {
    \draw[gray!50] (\centerx,\centery) -- ++(\angle:\radius);
}
\foreach \r/\label in {0.2/0.2, 0.4/0.4, 0.6/0.6, 0.8/0.8, 1.0/1.0} {
    \draw[gray!50] (\centerx,\centery) circle (\radius*\r);
    \node[anchor=west] at (90:\radius*\r) {\label};
}

\def\coleur{blue}
\foreach \idx in {0, 1, ..., 10} {
    \pgfmathsetmacro{\angle}{\anglesmat[\idx]}   
    \pgfmathsetmacro{\datapoint}{\datarowA[\idx]} 
    \coordinate (\idx) at (\angle:\radius*\datapoint); 
}
\draw[thick,\coleur] (0) -- (1) -- (2) -- (3) -- (4) -- (5) -- (6) -- (7) -- (8) -- (9) -- (10) -- cycle;
\foreach \point in {0, 1, ..., 10}  {
    \fill[\coleur] (\point) circle (2pt);
}

\def\coleur{brown}
\foreach \idx in {0, 1, ..., 10} {
    \pgfmathsetmacro{\angle}{\anglesmat[\idx]}   
    \pgfmathsetmacro{\datapoint}{\datarowB[\idx]} 
    \coordinate (\idx) at (\angle:\radius*\datapoint); 
}
\draw[thick,\coleur] (0) -- (1) -- (2) -- (3) -- (4) -- (5) -- (6) -- (7) -- (8) -- (9) -- (10) -- cycle;
\foreach \point in {0, 1, ..., 10}  {
    \fill[\coleur] (\point) circle (2pt);
}

\def\coleur{purple}
\foreach \idx in {0, 1, ..., 10} {
    \pgfmathsetmacro{\angle}{\anglesmat[\idx]}   
    \pgfmathsetmacro{\datapoint}{\datarowC[\idx]} 
    \coordinate (\idx) at (\angle:\radius*\datapoint); 
}
\draw[thick,\coleur] (0) -- (1) -- (2) -- (3) -- (4) -- (5) -- (6) -- (7) -- (8) -- (9) -- (10) -- cycle;
\foreach \point in {0, 1, ..., 10}  {
    \fill[\coleur] (\point) circle (2pt);
}

\def\coleur{yellow}
\foreach \idx in {0, 1, ..., 10} {
    \pgfmathsetmacro{\angle}{\anglesmat[\idx]}   
    \pgfmathsetmacro{\datapoint}{\datarowD[\idx]} 
    \coordinate (\idx) at (\angle:\radius*\datapoint); 
}
\draw[thick,\coleur] (0) -- (1) -- (2) -- (3) -- (4) -- (5) -- (6) -- (7) -- (8) -- (9) -- (10) -- cycle;
\foreach \point in {0, 1, ..., 10}  {
    \fill[\coleur] (\point) circle (2pt);
}

\def\coleur{orange}
\foreach \idx in {0, 1, ..., 10} {
    \pgfmathsetmacro{\angle}{\anglesmat[\idx]}   
    \pgfmathsetmacro{\datapoint}{\datarowE[\idx]} 
    \coordinate (\idx) at (\angle:\radius*\datapoint); 
}
\draw[thick,\coleur] (0) -- (1) -- (2) -- (3) -- (4) -- (5) -- (6) -- (7) -- (8) -- (9) -- (10) -- cycle;
\foreach \point in {0, 1, ..., 10}  {
    \fill[\coleur] (\point) circle (2pt);
}

\def\coleur{red}
\foreach \idx in {0, 1, ..., 10} {
    \pgfmathsetmacro{\angle}{\anglesmat[\idx]}   
    \pgfmathsetmacro{\datapoint}{\datarowF[\idx]} 
    \coordinate (\idx) at (\angle:\radius*\datapoint); 
}
\draw[thick,\coleur] (0) -- (1) -- (2) -- (3) -- (4) -- (5) -- (6) -- (7) -- (8) -- (9) -- (10) -- cycle;
\foreach \point in {0, 1, ..., 10}  {
    \fill[\coleur] (\point) circle (2pt);
}

\foreach \angle/\label in {0/$F_R$, 32.73/$F_G$, 65.46/$G_F$, 98.19/$C_L$, 130.92/$T_D$, 163.65/$L_N$, 196.38/$N_D$, 229.11/$S_C$, 261.84/$U_L$, 294.57/$S_W$, 327.30/$I_W$} {
    \node at (\angle:\radius*1.12) {\label};
}
\end{tikzpicture}
\caption{LLAMA-DPO}
\end{subfigure}

\begin{subfigure}[b]{0.45\textwidth} 
\centering
\begin{tikzpicture}
\def\centerx{0}
\def\centery{0}
\def\radius{2}
\def\angles{0, 32.73, 65.46, 98.19, 130.92, 163.65, 196.38, 229.11, 261.84, 294.57, 327.30}
\def\anglesmat{{\angles}}

\foreach \angle in \angles {
    \draw[gray!50] (\centerx,\centery) -- ++(\angle:\radius);
}
\foreach \r/\label in {0.2/0.2, 0.4/0.4, 0.6/0.6, 0.8/0.8, 1.0/1.0} {
    \draw[gray!50] (\centerx,\centery) circle (\radius*\r);
    \node[anchor=west] at (90:\radius*\r) {\label};
}

\def\datarowA{{0.2136, 0.0422, 0.0142, 0.0464, 0.0413, 0.0365, 0.1309, 0.0603, 0.0206, 0.1829, 0.0846}}
\def\datarowB{{0.3061, 0.145, 0.1216, 0.2495, 0.2475, 0.1659, 0.3218, 0.2056, 0.2019, 0.2829, 0.1615}}
\def\datarowC{{0.4364, 0.2683, 0.2604, 0.4132, 0.4463, 0.2976, 0.4727, 0.3553, 0.385, 0.44, 0.3077}}
\def\datarowD{{0.6001, 0.4372, 0.4184, 0.6095, 0.5913, 0.5235, 0.6955, 0.5847, 0.6072, 0.6086, 0.4538}}
\def\datarowE{{0.8037, 0.6279, 0.5856, 0.785, 0.6975, 0.72, 0.8564, 0.7797, 0.8134, 0.7714, 0.6692}}
\def\datarowF{{0.998, 0.8059, 0.7386, 0.9191, 0.7463, 0.8853, 0.9673, 0.9269, 1.0, 0.92, 1.0}}

\def\coleur{blue}
\foreach \idx in {0, 1, ..., 10} {
    \pgfmathsetmacro{\angle}{\anglesmat[\idx]}   
    \pgfmathsetmacro{\datapoint}{\datarowA[\idx]} 
    \coordinate (\idx) at (\angle:\radius*\datapoint); 
}
\draw[thick,\coleur] (0) -- (1) -- (2) -- (3) -- (4) -- (5) -- (6) -- (7) -- (8) -- (9) -- (10) -- cycle;
\foreach \point in {0, 1, ..., 10}  {
    \fill[\coleur] (\point) circle (2pt);
}

\def\coleur{brown}
\foreach \idx in {0, 1, ..., 10} {
    \pgfmathsetmacro{\angle}{\anglesmat[\idx]}   
    \pgfmathsetmacro{\datapoint}{\datarowB[\idx]} 
    \coordinate (\idx) at (\angle:\radius*\datapoint); 
}
\draw[thick,\coleur] (0) -- (1) -- (2) -- (3) -- (4) -- (5) -- (6) -- (7) -- (8) -- (9) -- (10) -- cycle;
\foreach \point in {0, 1, ..., 10}  {
    \fill[\coleur] (\point) circle (2pt);
}

\def\coleur{purple}
\foreach \idx in {0, 1, ..., 10} {
    \pgfmathsetmacro{\angle}{\anglesmat[\idx]}   
    \pgfmathsetmacro{\datapoint}{\datarowC[\idx]} 
    \coordinate (\idx) at (\angle:\radius*\datapoint); 
}
\draw[thick,\coleur] (0) -- (1) -- (2) -- (3) -- (4) -- (5) -- (6) -- (7) -- (8) -- (9) -- (10) -- cycle;
\foreach \point in {0, 1, ..., 10}  {
    \fill[\coleur] (\point) circle (2pt);
}

\def\coleur{yellow}
\foreach \idx in {0, 1, ..., 10} {
    \pgfmathsetmacro{\angle}{\anglesmat[\idx]}   
    \pgfmathsetmacro{\datapoint}{\datarowD[\idx]} 
    \coordinate (\idx) at (\angle:\radius*\datapoint); 
}
\draw[thick,\coleur] (0) -- (1) -- (2) -- (3) -- (4) -- (5) -- (6) -- (7) -- (8) -- (9) -- (10) -- cycle;
\foreach \point in {0, 1, ..., 10}  {
    \fill[\coleur] (\point) circle (2pt);
}

\def\coleur{orange}
\foreach \idx in {0, 1, ..., 10} {
    \pgfmathsetmacro{\angle}{\anglesmat[\idx]}   
    \pgfmathsetmacro{\datapoint}{\datarowE[\idx]} 
    \coordinate (\idx) at (\angle:\radius*\datapoint); 
}
\draw[thick,\coleur] (0) -- (1) -- (2) -- (3) -- (4) -- (5) -- (6) -- (7) -- (8) -- (9) -- (10) -- cycle;
\foreach \point in {0, 1, ..., 10}  {
    \fill[\coleur] (\point) circle (2pt);
}

\def\coleur{red}
\foreach \idx in {0, 1, ..., 10} {
    \pgfmathsetmacro{\angle}{\anglesmat[\idx]}   
    \pgfmathsetmacro{\datapoint}{\datarowF[\idx]} 
    \coordinate (\idx) at (\angle:\radius*\datapoint); 
}
\draw[thick,\coleur] (0) -- (1) -- (2) -- (3) -- (4) -- (5) -- (6) -- (7) -- (8) -- (9) -- (10) -- cycle;
\foreach \point in {0, 1, ..., 10}  {
    \fill[\coleur] (\point) circle (2pt);
}

\foreach \angle/\label in {0/$F_R$, 32.73/$F_G$, 65.46/$G_F$, 98.19/$C_L$, 130.92/$T_D$, 163.65/$L_N$, 196.38/$N_D$, 229.11/$S_C$, 261.84/$U_L$, 294.57/$S_W$, 327.30/$I_W$} {
    \node at (\angle:\radius*1.12) {\label};
}
\end{tikzpicture}
\caption{QWEN-FT}
\end{subfigure}
\hfill
\begin{subfigure}[b]{0.45\textwidth} 
\centering
\begin{tikzpicture}
\def\centerx{0}
\def\centery{0}
\def\radius{2}
\def\angles{0, 32.73, 65.46, 98.19, 130.92, 163.65, 196.38, 229.11, 261.84, 294.57, 327.30}
\def\anglesmat{{\angles}}

\foreach \angle in \angles {
    \draw[gray!50] (\centerx,\centery) -- ++(\angle:\radius);
}
\foreach \r/\label in {0.2/0.2, 0.4/0.4, 0.6/0.6, 0.8/0.8, 1.0/1.0} {
    \draw[gray!50] (\centerx,\centery) circle (\radius*\r);
    \node[anchor=west] at (90:\radius*\r) {\label};
}

\def\datarowA{{0.103, 0.0, 0.0, 0.0, 0.016, 0.011, 0.093, 0.037, 0.024, 0.151, 0.077}}
\def\datarowB{{0.260, 0.105, 0.141, 0.207, 0.208, 0.128, 0.266, 0.171, 0.184, 0.263, 0.162}}
\def\datarowC{{0.449, 0.276, 0.314, 0.438, 0.436, 0.27, 0.426, 0.332, 0.366, 0.449, 0.315}}
\def\datarowD{{0.623, 0.478, 0.490, 0.659, 0.643, 0.51, 0.684, 0.58, 0.573, 0.586, 0.477}}
\def\datarowE{{0.868, 0.699, 0.678, 0.843, 0.725, 0.729, 0.883, 0.788, 0.775, 0.797, 0.777}}
\def\datarowF{{1.0, 0.901, 0.842, 0.976, 0.778, 0.923, 1.0, 0.965, 1.0, 0.966, 1.0}}

\def\coleur{blue}
\foreach \idx in {0, 1, ..., 10} {
    \pgfmathsetmacro{\angle}{\anglesmat[\idx]}   
    \pgfmathsetmacro{\datapoint}{\datarowA[\idx]} 
    \coordinate (\idx) at (\angle:\radius*\datapoint); 
}
\draw[thick,\coleur] (0) -- (1) -- (2) -- (3) -- (4) -- (5) -- (6) -- (7) -- (8) -- (9) -- (10) -- cycle;
\foreach \point in {0, 1, ..., 10}  {
    \fill[\coleur] (\point) circle (2pt);
}

\def\coleur{brown}
\foreach \idx in {0, 1, ..., 10} {
    \pgfmathsetmacro{\angle}{\anglesmat[\idx]}   
    \pgfmathsetmacro{\datapoint}{\datarowB[\idx]} 
    \coordinate (\idx) at (\angle:\radius*\datapoint); 
}
\draw[thick,\coleur] (0) -- (1) -- (2) -- (3) -- (4) -- (5) -- (6) -- (7) -- (8) -- (9) -- (10) -- cycle;
\foreach \point in {0, 1, ..., 10}  {
    \fill[\coleur] (\point) circle (2pt);
}

\def\coleur{purple}
\foreach \idx in {0, 1, ..., 10} {
    \pgfmathsetmacro{\angle}{\anglesmat[\idx]}   
    \pgfmathsetmacro{\datapoint}{\datarowC[\idx]} 
    \coordinate (\idx) at (\angle:\radius*\datapoint); 
}
\draw[thick,\coleur] (0) -- (1) -- (2) -- (3) -- (4) -- (5) -- (6) -- (7) -- (8) -- (9) -- (10) -- cycle;
\foreach \point in {0, 1, ..., 10}  {
    \fill[\coleur] (\point) circle (2pt);
}

\def\coleur{yellow}
\foreach \idx in {0, 1, ..., 10} {
    \pgfmathsetmacro{\angle}{\anglesmat[\idx]}   
    \pgfmathsetmacro{\datapoint}{\datarowD[\idx]} 
    \coordinate (\idx) at (\angle:\radius*\datapoint); 
}
\draw[thick,\coleur] (0) -- (1) -- (2) -- (3) -- (4) -- (5) -- (6) -- (7) -- (8) -- (9) -- (10) -- cycle;
\foreach \point in {0, 1, ..., 10}  {
    \fill[\coleur] (\point) circle (2pt);
}

\def\coleur{orange}
\foreach \idx in {0, 1, ..., 10} {
    \pgfmathsetmacro{\angle}{\anglesmat[\idx]}   
    \pgfmathsetmacro{\datapoint}{\datarowE[\idx]} 
    \coordinate (\idx) at (\angle:\radius*\datapoint); 
}
\draw[thick,\coleur] (0) -- (1) -- (2) -- (3) -- (4) -- (5) -- (6) -- (7) -- (8) -- (9) -- (10) -- cycle;
\foreach \point in {0, 1, ..., 10}  {
    \fill[\coleur] (\point) circle (2pt);
}

\def\coleur{red}
\foreach \idx in {0, 1, ..., 10} {
    \pgfmathsetmacro{\angle}{\anglesmat[\idx]}   
    \pgfmathsetmacro{\datapoint}{\datarowF[\idx]} 
    \coordinate (\idx) at (\angle:\radius*\datapoint); 
}
\draw[thick,\coleur] (0) -- (1) -- (2) -- (3) -- (4) -- (5) -- (6) -- (7) -- (8) -- (9) -- (10) -- cycle;
\foreach \point in {0, 1, ..., 10}  {
    \fill[\coleur] (\point) circle (2pt);
}

\foreach \angle/\label in {0/$F_R$, 32.73/$F_G$, 65.46/$G_F$, 98.19/$C_L$, 130.92/$T_D$, 163.65/$L_N$, 196.38/$N_D$, 229.11/$S_C$, 261.84/$U_L$, 294.57/$S_W$, 327.30/$I_W$} {
    \node at (\angle:\radius*1.12) {\label};
}
\end{tikzpicture}
\caption{QWEN-DPO}
\end{subfigure}

\fbox{
\small{
    \begin{tabular}{llllllllllll}
        \textcolor{blue}{\rule{0.5cm}{2pt}} & $t=0.0$ &
        \textcolor{brown}{\rule{0.5cm}{2pt}} & $t=0.2$ &
        \textcolor{purple}{\rule{0.5cm}{2pt}} & $t=0.4$ &
        \textcolor{yellow}{\rule{0.5cm}{2pt}} & $t=0.6$ &
        \textcolor{orange}{\rule{0.5cm}{2pt}} & $t=0.8$ &
        \textcolor{red}{\rule{0.5cm}{2pt}} & $t=1.0$ 
    \end{tabular}
    }}

\caption{Comparison of linguistic feature controllability across models for varying target language proficiency ($t$). Each subfigure presents the distribution of 11 linguistic features at different target complexity levels: $t = 0.0, 0.2, 0.4, 0.6, 0.8, 1.0$. The concentric circles represent increasing values of each feature. Colors indicate the actual values achieved by the model for each feature under the specified target configuration. A greater deviation of the colored line from the corresponding circle indicates weaker control over that feature.}
\label{fig:ling-comparison}
\end{figure*}

The analysis of Figure~\ref{fig:ling-comparison} reveals that the QWEN and LLAMA models demonstrate comparable controllability across linguistic features, with different features exhibiting distinct trends. Among these, Utterance Length ($U_L$) is the most accurately controlled feature, consistently aligning with target complexity levels. This is likely because $U_L$ is a surface-form feature, which is comparatively easier for the model to manipulate. In contrast, Tree Depth ($T_D$) demonstrates strong control at lower complexity levels but weaker control at higher levels ($t=0.8, 1.0$). Conversely, Non-terminal Diversity ($N_D$) and Flesch Reading Ease ($F_R$) exhibit stronger controllability at higher complexity levels but struggle at lower levels. Subtree Complexity ($S_C$) and Simple Word Ratio ($S_W$) show moderate consistency across the range of complexity levels.

\subsection{Ablation Study on Linguistic Features}
To investigate the impact of eleven linguistic features, we conducted experiments by removing each feature individually. As summarized in Table~\ref{tab:ablation-study-on-single-feature}, the results demonstrate that removing most features, except for Utterance Length ($U_L$), negatively affects model performance, leading to declines in both AUC and DR. Notably, the removal of the Simple Word Ratio ($S_W$) results in the most significant performance drop. Conversely, removing Utterance Length improves all metrics, suggesting that constraints on utterance length may adversely affect dialogue quality. For instance, conflicts may arise when the dialogue task remains incomplete due to reaching the utterance length limit. Particularly, as shown in Figure~\ref{fig:ling-comparison}, Utterance Length is a feature over which the model exhibits strong control, indicating that the model tends to adapt to fit the constraint imposed by Utterance Length. A more detailed analysis is provided in Appendix~\ref{sec:analysis-on-ul}.

\section{Conclusion}
In this work, we propose leveraging three categories of linguistic features to regulate the language proficiency level of text generated by LLMs. Our experiments demonstrate that dialogue models trained on linguistically annotated dialogue data exhibit superior controllability over language proficiency. Specifically, these models achieve a wider controlling range of language proficiency, and enhanced stability in generating language at varying levels of difficulty, while simultaneously maintaining better overall dialogue quality.  Although setting $\tau(x_i)$ to a uniform  value for all features during inference demonstrates strong performance, our approach offers the flexibility to adjust individual features with different $\tau(x_i)$ values.  For instance, it allows for increasing grammatical complexity while keeping vocabulary simple, or vice versa, thereby enabling fine-grained control over the linguistic characteristics of the generated text. To comprehensively evaluate dialogue language proficiency, we introduce the Dilaprix metric, which integrates readability, syntactic, and lexical features into a unified measure. We demonstrate that Dilaprix exhibits a strong correlation with expert judgments, indicating its reliability as a metric for assessing dialogue language difficulty.

\section*{Limitations}
While training LLMs with linguistic features offers an effective means of controlling language difficulty, this approach requires substantial training data, which is not always readily available or feasible to acquire. Our research indicates that Dilaprix is a reliable index for quantifying the difficulty of dialogue. However, Dilaprix is not directly transferable to other text types, such as essays or reading materials without modification. Specifically, feature computation methods may require adjustment; for example, a feature such as Tree Depth, which is calculated as the maximum value across all sentences for dialogues, might be more appropriately represented by an average value when applied to other text types like essays or reading materials. Another limitation is the AUC metric's equal weighting of Dilaprix and RSR. This may not always be appropriate, as attributes like dialogue naturalness and consistency can be more important than the precise control of language proficiency.

\bibliography{custom}

\appendix

\section{Textbook Dialogue}
\label{sec:workflow-textbook-dialog}

Textbook dialogue is a specialized dialogue task tailored for early-stage L2 learners, who often encounter challenges in open-ended conversations and benefit from structured guidance. This task is organized around a series of predefined dialogue tasks curated by language teaching experts. The design aims to help students practice speaking skills in alignment with their textbooks, ensuring that the dialogue flow remains pedagogically relevant and focused.

To support this task, we generated 1,200 dialogue topics, curated by language teaching experts, based on various versions of K12 English textbooks.  A typical textbook dialogue topic consists of four to eight tasks, each designed to address specific language learning objectives. In each dialogue session, only one topic is discussed, and the session concludes either when all tasks within the topic have been addressed or when the user explicitly ends the dialogue. Throughout the session, the LLM is instructed to maintain the dialogue within the scope of the predefined topic, ensuring that the conversation remains focused on the specified tasks.

As the example topic shown in Figure~\ref{fig:dialogue-task-example}, a typical textbook dialogue consists of  three key components: the background, the AI's tasks, and the user's tasks. The background provides a brief description of the scene and the characters involved, setting the context for both the AI and the user. The AI's tasks usually include four to eight predefined tasks that LLM is required to follow during the conversation. The user's tasks are not directly used in this work, but they are presented to students as suggested responses to guide their participation. After the conversation concludes, the student's replies are evaluated based on the user's tasks. 

Throughout the conversation, an LLM-powered dialogue management agent, controls the conversation flow and determines the specific dialogue task to address at each turn. Based on this, prompts similar to the example presented in Figure~\ref{fig:prompt-example} are formulated for the dialogue model to generate appropriate responses.

\begin{figure}
    \centering
   
    \begin{tcolorbox}[colback=gray!1!white,colframe=black!75!black,arc=3mm]
    \footnotesize

\textbf{Background} \\
AI plays the role of Lily and User plays the role of Tom. In the classroom, Lily talks to her classmate Tom about their new teacher.
\vspace{10 pt}

\textbf{Lily's Tasks} \\
    1. Greet Tom and ask how he is.\par
    2. Say that you are fine and ask Tom what the new teacher’s name.\par
    3. Ask Tom if he likes his new teacher.\par
    4. Say goodbye to Tom.\par
\vspace{10 pt}
\textbf{Tom's Tasks} \\
    1. Say that you are good and ask Lily how she is.\par
    2. Say that the new teacher is Miss Lee.\par
    3. Say that you like your new teacher and she is very nice.\par
    4. Say goodbye to Lily.\par

\end{tcolorbox}
    \caption{An example of textbook dialogue topic.}
    \label{fig:dialogue-task-example}
\end{figure}

\section{Dialogue Quality Assessment}
\label{sec:dialog-assess}

Automatic dialogue evaluation leveraging LLMs has become an increasingly common approach. In this work, we employ \textit{Qwen-max} as the evaluator. Within the context of textbook dialogues, a response is deemed correct if the dialogue model both (1) responds naturally to the user's utterance and (2) accurately executes the specified task. The prompt designed for this evaluation, detailed in Figure~\ref{fig:prompt-assess}, incorporates established prompting techniques such as in-context learning and chain-of-thought.

To validate the performance of our textbook dialogue evaluator, we constructed a benchmark dataset comprising 738 samples, balanced between positive and negative instances. As presented in Table~\ref{tab:result-dialogue-evaluator}, the evaluator achieved an accuracy of 0.977, alongside nearly balanced precision (0.975) and recall (0.978) scores. These results underscore the reliability of the proposed dialogue evaluator.

\begin{figure*}
\begin{tcolorbox}[colback=gray!1!white,colframe=black!75!black,arc=3mm]
\footnotesize
You are required to evaluate whether the final expression in the dialogue meets the requirements set by the user task.
 \vspace{4 pt} 
 
\# Evaluation Method\par
1. Carefully review the examples provided above to understand the reasons for reasonableness or unreasonableness.\par
2. Compare the examples and evaluation criteria to ensure a clear understanding of the standards.\par
3. Analyze each criterion in detail to determine compliance before providing a final conclusion. Ensure the final conclusion strictly follows the required format.\par
\vspace{4 pt} 

\# Unreasonable Situations in \{role\_a\}'s Expression\par
\vspace{4 pt} 
1. Addressing Issues: \{role\_a\}'s expression is directed at \{role\_b\}. Pay special attention to whether names mentioned in the dialogue conform to conversational norms. Common issues include: Using the other person's name when second-person pronouns should be used.\par
2. Commentary Statements: \{role\_a\}'s expression should be from a first-person perspective; any additional narration or commentary statements are considered unreasonable.\par
3. Failure to Execute Task or Incomplete Task Execution: \{role\_a\}'s expression should include all elements of the required task. Common problems include:\par
 a). The task requires asking a question, but \{role\_a\} either failed to ask or provided an answer along with the question.\par
 b). The task requires expressing multiple sentences, but \{role\_a\} only expressed part of the required content.\par
4. Repeating Previously Stated Content: Unless the task specifically requires \{role\_a\} to repeat earlier content (usually due to the system determining that \{role\_b\} went off-topic), such repetition is considered unreasonable.\par
5. Illogical Expression: For example, asking a question despite already knowing the answer.\par
\vspace{4 pt}
\# Output Field Requirements\par
Reasoning: [Reasoning process]\par
Final Conclusion: [Yes/No]\par
\vspace{4 pt}
\#  Here are four examples:
\vspace{4 pt}

Example 1:\par
[Dialogue]
Jane: Hello Lily, could you please come to the kitchen for a moment?
Lily: sure
Jane: Of course, dear. Now that you're here, I wanted to ask if you like meat?
Lily: what is meat
Jane: Meat is food from animals, like chicken, beef, or pork. Do you like eating meat, Lily?

[User Task]
Jane: Do you like eating meat\par

[Evaluation Result]
Reasoning: Jane completed the task. Although she repeated the content of the task, it was reasonable because Lily went off-topic.\par
Final Conclusion: Yes
\vspace{4 pt}

Example 2:\par
[Dialogue]
Jane: Hello Lily, could you please come to the kitchen for a moment?
Lily: sure
Jane: Hi Jane. Now that you're here, I wanted to ask if Lily like meat?

[User Task]
Jane: Do you like eating meat\par

[Evaluation Result]
Reasoning: Jane greeted herself as "Hi Jane" instead of saying "Hi Lily." Additionally, the phrase "if Lily like meat" is unnatural and should be "if you like meat." Thus, there is an issue with addressing.\par
Final Conclusion: No
\vspace{4 pt}

Example 3:\par
[Dialogue]
Jane: Hello Lily, could you please come to the kitchen for a moment?
Lily: sure
Jane: OK, here is Jane's response: Of course, dear. Now that you’re here, I wanted to ask if you like meat?

[User Task]
Jane: Do you like eating meat\par

[Evaluation Result]
Reasoning: The phrase "OK, here is Jane's response" is an unnecessary commentary statement and thus unreasonable.\par
Final Conclusion: No
\vspace{4 pt}

Example 4:\par
[Dialogue]
Jane: Hello Lily, could you please come to the kitchen for a moment?
Lily: sure
Jane: Of course, dear. Now that you're here, I wanted to ask if you like meat?
Lily: what is meat 
Jane: Now that you're here, dear. Meat is food from animals, like chicken, beef, or pork. Do you like eating meat, Lily?

[User Task]
Jane: Do you like eating meat\par

[Evaluation Result]
Reasoning: Jane completed the task. However, she repeated the introductory phrase 'Now that you're here' unnecessarily, which was not required.\par
Final Conclusion: No
\vspace{4 pt}

\# Dialogue to Be Evaluated:\par
[Dialogue]
\{dialogue\_context\}

[User Task]
\{role\_a\}: \{dialogue\_task\}

[Evaluation Result]\par

Reasoning: \par

Final Conclusion:

\end{tcolorbox}
\caption{Prompt for textbook dialogue evaluation. } 
\label{fig:prompt-assess}
\end{figure*}

\begin{figure*}
    \centering
\footnotesize
\begin{subfigure}[b]{0.45\textwidth} 
\centering
\begin{tikzpicture}
\def\centerx{0}
\def\centery{0}
\def\radius{2}
\def\angles{0, 32.73, 65.46, 98.19, 130.92, 163.65, 196.38, 229.11, 261.84, 294.57, 327.30}
\def\anglesmat{{\angles}}

\foreach \angle in \angles {
    \draw[gray!50] (\centerx,\centery) -- ++(\angle:\radius);
}
\foreach \r/\label in {0.2/0.2, 0.4/0.4, 0.6/0.6, 0.8/0.8, 1.0/1.0} {
    \draw[gray!50] (\centerx,\centery) circle (\radius*\r);
    \node[anchor=west] at (90:\radius*\r) {\label};
}

\def\datarowA{{0.103, 0.0, 0.0, 0.0, 0.016, 0.011, 0.093, 0.037, 0.024, 0.151, 0.077}}
\def\datarowB{{0.260, 0.105, 0.141, 0.207, 0.208, 0.128, 0.266, 0.171, 0.184, 0.263, 0.162}}
\def\datarowC{{0.449, 0.276, 0.314, 0.438, 0.436, 0.27, 0.426, 0.332, 0.366, 0.449, 0.315}}
\def\datarowD{{0.623, 0.478, 0.490, 0.659, 0.643, 0.51, 0.684, 0.58, 0.573, 0.586, 0.477}}
\def\datarowE{{0.868, 0.699, 0.678, 0.843, 0.725, 0.729, 0.883, 0.788, 0.775, 0.797, 0.777}}
\def\datarowF{{1.0, 0.901, 0.842, 0.976, 0.778, 0.923, 1.0, 0.965, 1.0, 0.966, 1.0}}

\def\coleur{blue}
\foreach \idx in {0, 1, ..., 10} {
    \pgfmathsetmacro{\angle}{\anglesmat[\idx]}   
    \pgfmathsetmacro{\datapoint}{\datarowA[\idx]} 
    \coordinate (\idx) at (\angle:\radius*\datapoint); 
}
\draw[thick,\coleur] (0) -- (1) -- (2) -- (3) -- (4) -- (5) -- (6) -- (7) -- (8) -- (9) -- (10) -- cycle;
\foreach \point in {0, 1, ..., 10}  {
    \fill[\coleur] (\point) circle (2pt);
}

\def\coleur{brown}
\foreach \idx in {0, 1, ..., 10} {
    \pgfmathsetmacro{\angle}{\anglesmat[\idx]}   
    \pgfmathsetmacro{\datapoint}{\datarowB[\idx]} 
    \coordinate (\idx) at (\angle:\radius*\datapoint); 
}
\draw[thick,\coleur] (0) -- (1) -- (2) -- (3) -- (4) -- (5) -- (6) -- (7) -- (8) -- (9) -- (10) -- cycle;
\foreach \point in {0, 1, ..., 10}  {
    \fill[\coleur] (\point) circle (2pt);
}

\def\coleur{purple}
\foreach \idx in {0, 1, ..., 10} {
    \pgfmathsetmacro{\angle}{\anglesmat[\idx]}   
    \pgfmathsetmacro{\datapoint}{\datarowC[\idx]} 
    \coordinate (\idx) at (\angle:\radius*\datapoint); 
}
\draw[thick,\coleur] (0) -- (1) -- (2) -- (3) -- (4) -- (5) -- (6) -- (7) -- (8) -- (9) -- (10) -- cycle;
\foreach \point in {0, 1, ..., 10}  {
    \fill[\coleur] (\point) circle (2pt);
}

\def\coleur{yellow}
\foreach \idx in {0, 1, ..., 10} {
    \pgfmathsetmacro{\angle}{\anglesmat[\idx]}   
    \pgfmathsetmacro{\datapoint}{\datarowD[\idx]} 
    \coordinate (\idx) at (\angle:\radius*\datapoint); 
}
\draw[thick,\coleur] (0) -- (1) -- (2) -- (3) -- (4) -- (5) -- (6) -- (7) -- (8) -- (9) -- (10) -- cycle;
\foreach \point in {0, 1, ..., 10}  {
    \fill[\coleur] (\point) circle (2pt);
}

\def\coleur{orange}
\foreach \idx in {0, 1, ..., 10} {
    \pgfmathsetmacro{\angle}{\anglesmat[\idx]}   
    \pgfmathsetmacro{\datapoint}{\datarowE[\idx]} 
    \coordinate (\idx) at (\angle:\radius*\datapoint); 
}
\draw[thick,\coleur] (0) -- (1) -- (2) -- (3) -- (4) -- (5) -- (6) -- (7) -- (8) -- (9) -- (10) -- cycle;
\foreach \point in {0, 1, ..., 10}  {
    \fill[\coleur] (\point) circle (2pt);
}

\def\coleur{red}
\foreach \idx in {0, 1, ..., 10} {
    \pgfmathsetmacro{\angle}{\anglesmat[\idx]}   
    \pgfmathsetmacro{\datapoint}{\datarowF[\idx]} 
    \coordinate (\idx) at (\angle:\radius*\datapoint); 
}
\draw[thick,\coleur] (0) -- (1) -- (2) -- (3) -- (4) -- (5) -- (6) -- (7) -- (8) -- (9) -- (10) -- cycle;
\foreach \point in {0, 1, ..., 10}  {
    \fill[\coleur] (\point) circle (2pt);
}

\foreach \angle/\label in {0/$F_R$, 32.73/$F_G$, 65.46/$G_F$, 98.19/$C_L$, 130.92/$T_D$, 163.65/$L_N$, 196.38/$N_D$, 229.11/$S_C$, 261.84/$U_L$, 294.57/$S_W$, 327.30/$I_W$} {
    \node at (\angle:\radius*1.12) {\label};
}
\end{tikzpicture}
\caption{QWEN-DPO}
\end{subfigure}
\hfill
\begin{subfigure}[b]{0.45\textwidth} 
\centering
\begin{tikzpicture}
\def\centerx{0}
\def\centery{0}
\def\radius{2}
\def\angles{0, 32.73, 65.46, 98.19, 130.92, 163.65, 196.38, 229.11, 261.84, 294.57, 327.30}
\def\anglesmat{{\angles}}

\foreach \angle in \angles {
    \draw[gray!50] (\centerx,\centery) -- ++(\angle:\radius);
}
\foreach \r/\label in {0.2/0.2, 0.4/0.4, 0.6/0.6, 0.8/0.8, 1.0/1.0} {
    \draw[gray!50] (\centerx,\centery) circle (\radius*\r);
    \node[anchor=west] at (90:\radius*\r) {\label};
}

\def\datarowA{{0.1415, 0.0, 0.0, 0.0138, 0.1075, 0.0112, 0.1127, 0.0516, 0.1719, 0.1971, 0.0846}}
\def\datarowB{{0.2528, 0.1153, 0.1094, 0.2485, 0.3525, 0.1865, 0.3691, 0.2419, 0.4006, 0.3286, 0.1308}}
\def\datarowC{{0.4335, 0.2831, 0.2827, 0.4507, 0.5288, 0.3406, 0.5364, 0.4066, 0.5731, 0.4971, 0.2154}}
\def\datarowD{{0.6604, 0.5114, 0.4985, 0.6874, 0.6562, 0.6029, 0.78, 0.6691, 0.8031, 0.66, 0.4154}}
\def\datarowE{{0.8821, 0.7386, 0.6991, 0.8688, 0.7913, 0.85, 0.9782, 0.9203, 1.0, 0.8057, 0.6615}}
\def\datarowF{{1.0, 0.9224, 0.8551, 0.9813, 0.9075, 1.0, 1.0, 1.0, 1.0, 0.9057, 0.8923}}

\def\coleur{blue}
\foreach \idx in {0, 1, ..., 10} {
    \pgfmathsetmacro{\angle}{\anglesmat[\idx]}   
    \pgfmathsetmacro{\datapoint}{\datarowA[\idx]} 
    \coordinate (\idx) at (\angle:\radius*\datapoint); 
}
\draw[thick,\coleur] (0) -- (1) -- (2) -- (3) -- (4) -- (5) -- (6) -- (7) -- (8) -- (9) -- (10) -- cycle;
\foreach \point in {0, 1, ..., 10}  {
    \fill[\coleur] (\point) circle (2pt);
}

\def\coleur{brown}
\foreach \idx in {0, 1, ..., 10} {
    \pgfmathsetmacro{\angle}{\anglesmat[\idx]}   
    \pgfmathsetmacro{\datapoint}{\datarowB[\idx]} 
    \coordinate (\idx) at (\angle:\radius*\datapoint); 
}
\draw[thick,\coleur] (0) -- (1) -- (2) -- (3) -- (4) -- (5) -- (6) -- (7) -- (8) -- (9) -- (10) -- cycle;
\foreach \point in {0, 1, ..., 10}  {
    \fill[\coleur] (\point) circle (2pt);
}

\def\coleur{purple}
\foreach \idx in {0, 1, ..., 10} {
    \pgfmathsetmacro{\angle}{\anglesmat[\idx]}   
    \pgfmathsetmacro{\datapoint}{\datarowC[\idx]} 
    \coordinate (\idx) at (\angle:\radius*\datapoint); 
}
\draw[thick,\coleur] (0) -- (1) -- (2) -- (3) -- (4) -- (5) -- (6) -- (7) -- (8) -- (9) -- (10) -- cycle;
\foreach \point in {0, 1, ..., 10}  {
    \fill[\coleur] (\point) circle (2pt);
}

\def\coleur{yellow}
\foreach \idx in {0, 1, ..., 10} {
    \pgfmathsetmacro{\angle}{\anglesmat[\idx]}   
    \pgfmathsetmacro{\datapoint}{\datarowD[\idx]} 
    \coordinate (\idx) at (\angle:\radius*\datapoint); 
}
\draw[thick,\coleur] (0) -- (1) -- (2) -- (3) -- (4) -- (5) -- (6) -- (7) -- (8) -- (9) -- (10) -- cycle;
\foreach \point in {0, 1, ..., 10}  {
    \fill[\coleur] (\point) circle (2pt);
}

\def\coleur{orange}
\foreach \idx in {0, 1, ..., 10} {
    \pgfmathsetmacro{\angle}{\anglesmat[\idx]}   
    \pgfmathsetmacro{\datapoint}{\datarowE[\idx]} 
    \coordinate (\idx) at (\angle:\radius*\datapoint); 
}
\draw[thick,\coleur] (0) -- (1) -- (2) -- (3) -- (4) -- (5) -- (6) -- (7) -- (8) -- (9) -- (10) -- cycle;
\foreach \point in {0, 1, ..., 10}  {
    \fill[\coleur] (\point) circle (2pt);
}

\def\coleur{red}
\foreach \idx in {0, 1, ..., 10} {
    \pgfmathsetmacro{\angle}{\anglesmat[\idx]}   
    \pgfmathsetmacro{\datapoint}{\datarowF[\idx]} 
    \coordinate (\idx) at (\angle:\radius*\datapoint); 
}
\draw[thick,\coleur] (0) -- (1) -- (2) -- (3) -- (4) -- (5) -- (6) -- (7) -- (8) -- (9) -- (10) -- cycle;
\foreach \point in {0, 1, ..., 10}  {
    \fill[\coleur] (\point) circle (2pt);
}

\foreach \angle/\label in {0/$F_R$, 32.73/$F_G$, 65.46/$G_F$, 98.19/$C_L$, 130.92/$T_D$, 163.65/$L_N$, 196.38/$N_D$, 229.11/$S_C$, 261.84/$U_L$, 294.57/$S_W$, 327.30/$I_W$} {
    \node at (\angle:\radius*1.12) {\label};
}
\end{tikzpicture}
\caption{QWEN-DPO-$U_L$}
\end{subfigure}

\fbox{
\small{
    \begin{tabular}{llllllllllll}
        \textcolor{blue}{\rule{0.5cm}{2pt}} & $t=0.0$ &
        \textcolor{brown}{\rule{0.5cm}{2pt}} & $t=0.2$ &
        \textcolor{purple}{\rule{0.5cm}{2pt}} & $t=0.4$ &
        \textcolor{yellow}{\rule{0.5cm}{2pt}} & $t=0.6$ &
        \textcolor{orange}{\rule{0.5cm}{2pt}} & $t=0.8$ &
        \textcolor{red}{\rule{0.5cm}{2pt}} & $t=1.0$ 
    \end{tabular}
    }}

\caption{Comparison of linguistic feature controllability between models with and without the Utterance Length ($U_L$) feature. The left subplot (QWEN-DPO) represents the model trained with $U_L$; the right subplot (QWEN-DPO-$U_L$) represents the model trained without $U_L$.}
\label{fig:ling-comparison-woul}
\end{figure*}

\begin{figure}
\footnotesize
\begin{tikzpicture}
\begin{axis}[
    xlabel={Dilaprix},
    ylabel={RSR},
    xmin=0, xmax=1,
    ymin=0.78, ymax=1,
    xtick={0,0.1,...,1},
    ytick={0.75,0.80,...,1},
    legend pos=south east,
    ymajorgrids=true,
    grid style=dashed,
    width=0.45\textwidth, 
    height=0.286\textwidth,
    legend style={font=\small, scale=0.8},
]


\addplot[
    color=orange,
    only marks,
    mark=*,
] coordinates {
(0.102,0.946) (0.176,0.957) (0.251,0.975) (0.33,0.99) (0.413,0.993) (0.53,0.993) (0.625,0.993) (0.716,0.988) (0.803,0.993) (0.863,0.988) (0.903,0.99)
};
\addlegendentry{QWEN-DPO-$U_L$}

\addplot[
    color=blue,
    only marks,
    mark=*,
] coordinates {
(0.073,0.872) (0.128,0.913) (0.192,0.953) (0.267,0.973) (0.361,0.99) (0.478,0.987) (0.565,0.977) (0.658,0.984) (0.752,0.977) (0.829,0.975) (0.883,0.963)
};
\addlegendentry{QWEN-DPO}

\addplot[no markers, blue, dashed,  mark=o, domain=0.073:0.883, samples=100] { -1.9422 * x^4 + 4.5878 * x^3 -4.0155 * x^2 + 1.5158 * x + 0.7791 };

\addplot[no markers, orange, dashed,  mark=o, domain=0.102:0.903, samples=100] { 9.8214 * x^6 -33.4598 * x^5 + 44.7439 * x^4 -29.2675 * x^3 + 9.3208 * x^2 -1.1505 * x + 0.9929 };


\end{axis}
\end{tikzpicture}
\caption{Comparison of dialogue quality (RSR) versus target language proficiency (Dilaprix) for models with and without the Utterance Length feature. Dashed lines represent fitted curves for each model.}
\label{fig:fig:dilaprix-rsr-line-ul} 
\end{figure}

\begin{table}[h]
\centering
\small
\begin{tabular}{cccc}
\toprule
\textbf{Accuracy} & \textbf{Precision}\ & \textbf{ Recall} & \textbf{ F1} \\ 
\midrule
\makecell{0.977\\ $\pm${0.0017}} & 
\makecell{0.975\\ $\pm${0.0012}} & 
\makecell{0.978\\ $\pm${0.0038}} & 
\makecell{0.976\\ $\pm${0.0018}} \\ 
\bottomrule
\end{tabular}
\caption{Performance of the Dialogue Evaluator. Results are reported as mean ± standard deviation from three experimental runs.}
\label{tab:result-dialogue-evaluator}
\end{table}

\section{Analysis on the Utterance Length}
\label{sec:analysis-on-ul}

\begin{table}[h]
\centering
\footnotesize
\begin{tabular}{l|ccc}
\toprule
\textbf{Method} & \textbf{AUC}\,($\uparrow$) & \textbf{ DR}\,($\uparrow$) & \textbf{ STD}\,($\downarrow$) \\ 
\midrule
QWEN-FT & 0.703 & 0.741 & 0.093 \\ 
- ${U_L}$ & 0.724 & 0.748 & 0.090 \\ 
\midrule
QWEN-DPO & {0.786} & {0.810} & {0.084} \\ 
- $U_L$ & {0.789} & {0.801} & {0.067} \\ 
\bottomrule
\end{tabular}
\caption{Performance comparison of language proficiency-controlled models with and without the Utterance Length feature. The `- $U_L$` rows represent models trained on data where the Utterance Length feature has been removed.}.
\label{tab:auc-results-ul}
\end{table}

We observe a significant performance improvement when fine-tuned on linguistic dialogue data with the Utterance Length feature removed from the controlling features (denoted as QWEN-FT-$U_L$) compared to the QWEN-FT model. We further optimize QWEN-FT-$U_L$ using DPO, resulting in QWEN-DPO-$U_L$. As the results summarized in Table~\ref{tab:auc-results-ul}, QWEN-FT-$U_L$ demonstrates improvements across all metrics compared to QWEN-FT, with a notable increase of 0.21 in AUC. However, the DPO narrows this improvement to 0.03 while reducing the Dilaprix range. The reduction in the Dilaprix range is not surprising, as relaxing the utterance length constraint increases the lower bound of Dilaprix. This subsequently decreases the overall controllability interval, when the Dilaprix upper bound does not correspondingly increase.

As shown in Figure~\ref{fig:fig:dilaprix-rsr-line-ul}, after relaxing the Utterance Length constraint, QWEN-DPO-$U_L$ achieves a significant increase in RSR when the Dilaprix is low (< 0.2). However, as the Dilaprix increases (> 0.5), the RSR of QWEN-DPO begins to decline gradually, whereas this trend is not observed for QWEN-DPO-$U_L$. This suggests that the Utterance Length constraint negatively impacts dialogue quality not only when Dilaprix is low but also when it is high.

As illustrated in Figure~\ref{fig:ling-comparison-woul}, the control behavior changes after removing the Utterance Length feature. Specifically, utterance lengths tend to increase across all levels of language proficiency, while the controllability of Tree Depth improves for high complexity levels ($t=0.8$, $t=1.0$) but decreases for lower complexity levels ($t=0.0$, $t=0.2$, $t=0.4$). Interestingly, the controllability of the Intermediate Word Ratio is diminished after removing the Utterance Length constraint.

\section{Experiment Details}
In this work, we utilized toolkits such as NLTK~\cite{bird-loper-2004-nltk} and spaCy~\cite{spacy2} for extracting linguistic features. Dialogue models were trained using LLaMA-Factory~\cite{zheng2024llamafactory}.  For fine-tuning, the training was conducted on the textbook dialogue dataset with a batch size of 64 for one epoch. We employed the Adam optimizer, setting $\beta_1=0.9, \beta_2=0.999$, a warmup ratio of $0.1$ and a learning rate of $1\times10^{-5}$. For DPO, we trained on  the textbook dialogue preference dataset with a batch size of 64 and a preference beta $\beta=0.1$ for one epoch.  Similarly, we used the Adam optimizer with $\beta_1=0.9, \beta_2=0.999$, a warmup ratio of $0.1$ and  a learning rate of $5\times10^{-7}$. The models were trained on eight A10 GPUs.

\begin{figure}
    \centering
   
    \begin{tcolorbox}[colback=gray!1!white,colframe=black!75!black,arc=3mm]
    \footnotesize

Compare the following two utterances from an English language teaching perspective based on  the criteria below, to determine which one is more difficult.\par

\vspace{4 pt}

Lexical Complexity\par
  -- Does the sentence include advanced vocabulary, technical terms, or rare words?\par
  -- Are there any polysemous or ambiguous words?\par

\vspace{4 pt}

Sentence Structure Complexity\par
  -- Does the sentence contain complex grammatical structures?\par
  -- Are there long sentences or multiple modifiers?\par

\vspace{4 pt}

Semantic Understanding Difficulty\par
  -- Is the meaning of the sentence clear and straightforward? \par
  -- Does it involve metaphors, similes, or other rhetorical devices? \par

\vspace{4 pt}

UTTERANCE A \par

\vspace{2 pt}

UTTERANCE B \par

\vspace{2 pt}

Which one is more difficult:

\end{tcolorbox}
    \caption{Instruction for language difficulty comparison.}
    \label{fig:instruction-for-comparison}
\end{figure}

\section{Human Annotation Instruction}
As described in Section~\ref{sec:dilaprix}, to validate the reliability of Dilaprix in assessing dialogue language proficiency, we conducted human evaluation. Given the inherent difficulty of directly assigning a difficulty score to individual utterances, we adopted a pairwise comparison approach. Figure~\ref{fig:instruction-for-comparison} presents the instructions provided to annotators. 

\section{CEFR Sentence Annotator}
To further validate the effectiveness of our approach, we employ a CEFR-based sentence difficulty annotator (CEFR-SP)~\cite{arase-etal-2022-cefr} for language difficulty evaluation. This tool predicts difficulty levels on a scale from 0 to 5, corresponding to the six CEFR levels ranging from easy to difficult. We first evaluate CEFR-SP on the dataset of 100 utterance samples with human annotations, and the CEFR-SP achieves a PCC of 0.839. Although CEFR-SP does not perform as well as Dilaprix in evaluating language difficulty, it remains a viable tool for assessing the difficulty of textbook dialogues.

As shown in Table \ref{tab:cefr-sp-results}, QWEN-DPO demonstrates a broader range of difficulty control compared to the Prompt-Based method (1.784 vs. 1.213) and achieves a lower average standard deviation of 0.181, as opposed to 0.272. These findings are consistent with the evaluation results obtained using Dilaprix.

\begin{table*}[h]
\centering
\small
\begin{tabular}{l|cccccc}
\toprule
\textbf{Method} &  \textbf{A1 ($t=0.0$)} & \textbf{A2 ($t=0.2$)} & \textbf{B1 ($t=0.4$)} & \textbf{B2 ($t=0.6$)} & \textbf{C1 ($t=0.8$)} & \textbf{C2 ($t=1.0$)} \\ 
\midrule
\multirow{1}{2.cm}
{Prompt-Based}  & \makecell{$1.082$  $\pm 0.243$} 
                & \makecell{$1.109$  $\pm 0.254$} 
                & \makecell{$1.596$  $\pm 0.279$} 
                & \makecell{$1.832$  $\pm 0.250$} 
                & \makecell{$1.839$  $\pm 0.218$} 
                & \makecell{$2.295$  $\pm 0.389$} \\
\midrule
{QWEN-DPO}      & \makecell{$0.685$  $\pm 0.229$} 
                & \makecell{$1.078$  $\pm 0.106$} 
                & \makecell{$1.588$  $\pm 0.251$} 
                & \makecell{$2.015$  $\pm 0.079$} 
                & \makecell{$2.197$  $\pm 0.160$} 
                & \makecell{$2.469$  $\pm 0.258$} \\
\bottomrule
\end{tabular}
\caption{Comparison of Language Difficulty Evaluation between the Prompt-Based Method and the QWEN-DPO Model using CEFR-SP. Columns A1--C2 represent CEFR levels for the Prompt-Based method, while $t \in \{0.0, 0.2, \dots, 1.0\}$ indicates difficulty levels for the QWEN-DPO model. }
\label{tab:cefr-sp-results}
\end{table*}

\begin{table*}[h]
\centering
\small
\begin{tabular}{l|cc|cc|cc}
\toprule
&  \textbf{Group1-A} & \textbf{Group1-B} & \textbf{Group2-A} & \textbf{Group2-B} & \textbf{Group3-A} & \textbf{Group3-B} \\ 
\midrule
\multirow{1}{1.5cm}
{Readability}  & \makecell{$0.257$  $\pm 0.019$} 
                & \makecell{$0.390$  $\pm 0.044$} 
                & \makecell{$0.555$  $\pm 0.023$} 
                & \makecell{$0.675$  $\pm 0.031$} 
                & \makecell{$0.336$  $\pm 0.017$} 
                & \makecell{$0.564$  $\pm 0.050$} \\
\midrule
{Lexical}       & \makecell{$0.264$  $\pm 0.032$} 
                & \makecell{$0.400$  $\pm 0.071$} 
                & \makecell{$0.493$  $\pm 0.041$} 
                & \makecell{$0.703$  $\pm 0.051$} 
                & \makecell{$0.298$  $\pm 0.026$} 
                & \makecell{$0.626$  $\pm 0.074$} \\
\midrule
{Syntactic   }  & \makecell{$0.454$  $\pm 0.011$} 
                & \makecell{$0.218$  $\pm 0.010$} 
                & \makecell{$0.736$  $\pm 0.010$} 
                & \makecell{$0.534$  $\pm 0.013$} 
                & \makecell{$0.686$  $\pm 0.012$} 
                & \makecell{$0.248$  $\pm 0.013$} \\
\midrule
{Dilaprix   }  & \makecell{$0.348$  $\pm 0.007$} 
                & \makecell{$0.314$  $\pm 0.014$} 
                & \makecell{$0.626$  $\pm 0.008$} 
                & \makecell{$0.616$  $\pm 0.011$} 
                & \makecell{$0.488$  $\pm 0.006$} 
                & \makecell{$0.432$  $\pm 0.014$} \\
\midrule
{RSR         }  & \makecell{$0.988$} 
                & \makecell{$0.950$} 
                & \makecell{$0.989$} 
                & \makecell{$0.967$} 
                & \makecell{$0.989$} 
                & \makecell{$0.915$} \\
\bottomrule
\end{tabular}
\caption{Results of the comparative experiments assessing the impact of varying $t$ for lexical/readability ($t_{\text{lex/read}}$) and syntactic ($t_{\text{syn}}$) features. The 'Readability', 'Lexical', and 'Syntactic' rows show the averaged normalized scores for their respective feature categories. Each group (1, 2, 3) compares a configuration (A) against its inverse (B)}
\label{tab:varying-t-experiments}
\end{table*}

\section{Effect of Varying $t$ for Lexical and Syntactic Features}
While employing a uniform $t$ for all features during inference yields strong performance, our approach allows for the granular adjustment of $t$ for different feature groups. This flexibility enables a more nuanced control over the model's behavior.

 The readability scores are composite metrics consisting of two components: a word complexity component (calculated based on the number of syllables or letters in words) and a sentence length component. An examination of the underlying formulas for the Flesch Reading Ease, Flesch-Kincaid Grade Level, and Coleman-Liau Index reveals that the lexical component is the predominant factor in the final score and the Gunning Fog Index shows a more balanced contribution of the lexical component and syntactic component.  Therefore, when considering our full suite of readability metrics, the collective influence of the lexical component is substantially greater than that of the syntactic component.  Consequently, in the following experiments, we group readability features with lexical features.

To assess the impact of assigning different $t$ values to the ``lexical/readability'' category versus the ``syntactic'' category, we designed three comparative experiments. The specific configurations for each experiment are detailed as following.

    \textbf{Group 1}. We compare the setting of $t_{\text{lex/read}} = 0.2$ and $t_{\text{syn}} = 0.5$ (Group1-A) against the inverse configuration of $t_{\text{lex/read}} = 0.5$ and $t_{\text{syn}} = 0.2$ (Group1-B).

    \textbf{Group 2}. We compare the setting of $t_{\text{lex/read}} = 0.5$ and $t_{\text{syn}} = 0.8$ (Group2-A) against the inverse configuration of $t_{\text{lex/read}} = 0.8$ and $t_{\text{syn}} = 0.5$ (Group2-B).

    \textbf{Group 3}. We compare the setting of $t_{\text{lex/read}} = 0.2$ and $t_{\text{syn}} = 0.8$ (Group3-A) against the inverse configuration of $t_{\text{lex/read}} = 0.8$ and $t_{\text{syn}} = 0.2$ (Group3-B).

Here, $t_{\text{lex/read}}$ is the value assigned to the lexical and readability features, while $t_{\text{syn}}$ corresponds to the syntactic features.

As the results shown in Table~\ref{tab:varying-t-experiments},  our method's ability to assign different $t$ values to feature categories provides a robust and predictable mechanism for fine-grained text attribute control. This flexibility is a significant advantage, enabling users to precisely tune model outputs to meet diverse requirements, such as generating text with simple vocabulary but complex sentence structures, or vice versa, while maintaining high overall quality. An interesting trend is observed in that for the B settings ($t_{\text{lex/read}} > t_{\text{syn}}$), the RSR is lower than for the A settings ($t_{\text{lex/read}} < t_{\text{syn}}$).

\end{document}